\documentclass[11pt,a4paper]{article}
\usepackage[hyperref]{acl2021}
\usepackage{times}
\usepackage{latexsym}

\usepackage{microtype}

\usepackage[T1]{fontenc}
\usepackage[utf8]{inputenc}
\usepackage{microtype}

\usepackage{graphicx}
\graphicspath{ {./images/}}
\usepackage{amsmath}
\usepackage{amssymb}
\usepackage{amsfonts}
\usepackage{amstext}
\usepackage{amsthm}
\usepackage{bbm}
\usepackage{arydshln}
\usepackage{tikz}
\usepackage{tikz-dependency}
\usepackage{algorithm}
\usepackage{algpseudocode}
\usepackage{qtree}
\usepackage{subcaption}
\usepackage[export]{adjustbox}
\usepackage{algorithmicx}
\usepackage{algpseudocode}

\definecolor{deepblue}{rgb}{0,0,0.5}
\definecolor{officeblue}{RGB}{0,102,204}
\definecolor{deepred}{rgb}{0.6,0,0}
\definecolor{deepgreen}{rgb}{0,0.5,0}
\definecolor{mylightgreen}{rgb}{0,128,0}
\definecolor{mybrickred}{RGB}{182,50,28}

\definecolor{fillcolor}{RGB}{216,217,252}

\algnewcommand\algorithmicrequireb{{\hspace{0.85cm}}}
\algnewcommand\INPTDESCB{\item[\algorithmicrequireb]}

\algnewcommand\algorithmicfuncdesc{\textbf{Function:}}
\algnewcommand\FUNCDESC{\item[\algorithmicfuncdesc]}
\algnewcommand\algorithmicfuncdescb{{\hspace{1.48cm}}}
\algnewcommand\FUNCDESCB{\item[\algorithmicfuncdescb]}
\algnewcommand{\algorithmicgoto}{\textbf{goto}}
\algnewcommand{\Goto}[1]{\algorithmicgoto~\ref{#1}}

\newcommand*\AlgCommentInLine[1]{{\color{deepblue}{$\triangleright$ \textit{#1}}}}
\newcommand*\AlgComment[1]{\State{\AlgCommentInLine{#1}}}
\newcommand{\calT}{\mathcal{T}}

\aclfinalcopy % Uncomment this line for the final submission
\title{Monotonicity Marking from Universal Dependency Trees}

\author{Zeming Chen \qquad Qiyue Gao \\
    Department of Computer Science and Software Engineering,\\ Rose-Hulman Institute of Technology\\

\tt{\{chenz16, gaoq\}@rose-hulman.edu}
}

\date{}

\begin{document}
\maketitle
\begin{abstract}
Dependency parsing is a tool widely used in the field of Natural Language Processing and computational linguistics. However, there is hardly any work that connects dependency parsing to monotonicity, which is an essential part of logic and linguistic semantics. In this paper, we present a system that automatically annotates monotonicity information based on Universal Dependency parse trees. Our system utilizes surface-level monotonicity facts about quantifiers, lexical items, and token-level polarity information. We compared our system's performance with existing systems in the literature, including NatLog and ccg2mono, on a small evaluation dataset. Results show that our system outperforms NatLog and ccg2mono.
\end{abstract}

\section{Introduction}
The number of computational approaches for Natural Language Inference (NLI) has rapidly grown in recent years. Most of the approaches can be categorized as (1) Systems that translate sentences into first-order logic expressions and then apply theorem proving \cite{Blackburn2005RepresentationAI}. (2) Systems that use blackbox neural network approaches to learn the inference \cite{devlin-etal-2019-bert,liu-etal-2019-multi}. (3) Systems that apply natural logic as a tool to make inferences \cite{maccartney-manning-2009-extended,monalog,angeli-etal-2016-combining, abzianidze-2017-langpro}. Compared to neural network approaches, systems that apply natural logic are more robust, formally more precise, and more explainable. Several systems contributed to the third category \cite{maccartney-manning-2009-extended,monalog,angeli-etal-2016-combining} to solve the NLI task using monotonicity reasoning, a type of logical inference that is based on word replacement. Below is an example of monotonicity reasoning:
 \begin{enumerate}
   \item
   \begin{enumerate}
     \item {\footnotesize \textbf{All} \underline{students}$\downarrow$ carry a \underline{MacBook}$\uparrow$.}
     \item {\footnotesize All students carry a \underline{laptop}.}
     \item {\footnotesize All \underline{new students} carry a MacBook.}
   \end{enumerate}
   \item
   \begin{enumerate}
     \item {\footnotesize \textbf{Not all} \underline{new students}$\uparrow$ carry a laptop.}
     \item {\footnotesize Not all \underline{students} carry a laptop.}
   \end{enumerate}
 \end{enumerate}

\noindent As the example shows, the word replacement is based on the polarity mark (arrow) on each word. A monotone polarity ($\uparrow$) allows an inference from (1a) to (1b), where a more general concept \textit{laptop} replaces the more specific concept \textit{MacBook}. An antitone polarity ($\downarrow$) allows an inference from (1a) to (1c), where a more specific concept \textit{new students} replaces the more general concept \textit{students}. The direction of the polarity marks can be reversed by adding a downward entailment operator like \textit{Not} which allows an inference from (2a) to (2b). Thus, successful word placement relies on accurate polarity marks. To obtain the polarity mark for each word, an automatic polarity marking system is required to annotate a sentence by placing polarity mark on each word. This is formally called the polarization process. Polarity markings support monotonicity reasoning, and thus are used by systems for Natural Language Inference and data augmentations for language models. \cite{maccartney-manning-2009-extended,monalog,angeli-etal-2016-combining}. 

In this paper, we introduce a novel automatic polarity marking system that annotates monotonicity information by applying a polarity algorithm on a universal dependency parse tree. Our system is inspired by ccg2mono, an automatic polarity marking system \cite{hu-moss-2018-polarity} used by \citet{monalog}. In contrast to ccg2mono, which derives monotonicity information from CCG \cite{lewis-steedman-2014-ccg}  parse trees, our system's polarization algorithm derives monotonicity information using Universal Dependency \cite{nivre-etal-2016-universal} parse trees. There are several advantages of using UD parsing for polarity marking rather than CCG parsing. First, UD parsing is more accurate since the amount of training data for UD parsing is larger than those of CCG parsing. The high accuracy of UD parsing should lead to more accurate polarity annotation. Second, UD parsing works for more types of text. Overall, our system opens up a new framework for performing inference, semantics, and automated reasoning over UD representations. We will introduce the polarization algorithm's general steps, a set of rules we used to mark polarity on dependency parse trees, and comparisons between our system and some existing polarity marking tools, including NatLog \cite{maccartney-manning-2009-extended, angeli-etal-2016-combining} and ccg2mono. Our evaluation focuses on a small dataset used to evaluate ccg2mono \cite{huMoss2020Tsinghua}. Our system outperforms NatLog and ccg2mono. In particular, our system achieves the highest annotation accuracy on both the token level and the sentence level.

\section{Related Work}
Universal Dependencies (UD) \cite{nivre-etal-2016-universal} was first designed to handle language tasks for many different languages. The syntactic annotation in UD mostly relies on dependency relations. Words enter into dependency relations, and that is what UD tries to capture. There are 40 grammatical dependency relations between words, such as nominal subject (\textbf{nsubj}), relative clause modifier (\textbf{acl:relcl}), and determiner (\textbf{det}). A dependency relation connects a headword to a modifier. For example, in the dependency parse tree for \textit{All dogs eat food} (figure 1), the dependency relation \textbf{nsubj} connects the modifier \textit{dogs} and the headword \textit{eat}. The system presented in this paper utilizes Universal Dependencies to obtain a dependency parse tree from a sentence. We will explain the details of the parsing process in the implementation section.

There are two relevant systems of prior work: (1) The NatLog \cite{maccartney-manning-2009-extended, angeli-etal-2016-combining} system included in the Stanford CoreNLP library \cite{manning-etal-2014-stanford}; (2) The ccg2mono system \cite{hu-moss-2018-polarity}. 
The NatLog system is a natural language inference system, a part of the Stanford CoreNLP Library. NatLog marks polarity to each sentence by applying a pattern-based polarization algorithm to the dependency parse tree generated by the Stanford dependency parser. A list of downward-monotone and non-monotone expressions are defined along with an arity and a Tregex pattern for the system to identify if an expression occurred. 

The ccg2mono system is a polarity marking tool that annotates a sentence by polarizing a CCG parse tree. The polarization algorithm of ccg2mono is based on \citet{vanBenthemEssays86}'s work and \citet{Moss2012TheSO}'s continuation on the soundness of internalized polarity marking. The system uses a marked/order-enriched lexicon and can handle application rules, type-raising, and composition in CCG. The main polarization contains two steps: mark and polarize. For the mark step, the system puts markings on each node in the parse tree
from leaf to root. For the polarize step, the system generates polarities to each node from root to leaf. Compared to NatLog, an advantage of ccg2mono is that it polarizes on both the word-level and the constituent level.

\begin{figure}[t]
    \centering
    \includegraphics[width=5cm]{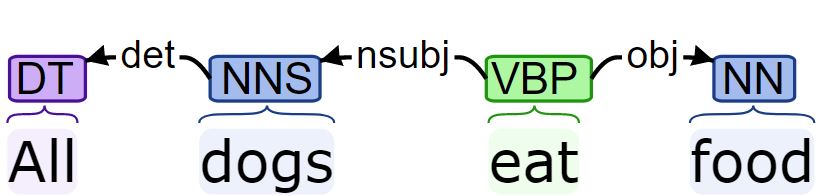}
    \caption{A dependency parse tree for "All dogs eat food."}
    \label{fig:1}
\end{figure}

\section{Universal Dependency to Polarity}
\subsection{Overview}
Our system's polarization algorithm contains three steps: (1) Universal Dependency Parsing, which transforms a sentence to a UD parse tree, (2) Binarization, which converts a UD parse tree to a binary UD parse tree, and (3) Polarization, which places polarity marks on each node in a binary UD parse tree.

\begin{table}[!b]
    \centering
    \small
    \begin{tabular}{|c|c|c|c|}
        \hline relation & level-id & relation & level-id \\ \hline
            conj-sent & 0       &   obl:tmod & 50 \\
            advcl-sent & 1      &   obl:npmod & 50 \\
            advmod-sent & 2     &   cop & 50 \\
            case & 10           &   det & 55 \\
            mark & 10           &   det:predet & 55 \\
            expl & 10           &   acl & 60 \\
            discourse & 10      &   acl:relcl & 60 \\
            nsubj & 20          &   appos & 60 \\
            csubj & 20          &   conj & 60 \\
            nsubj:pass & 20     &   conj-np & 60 \\
            conj-vp & 25        &   conj-adj & 60 \\
            ccomp & 30          &   obj & 60 \\
            advcl & 30          &   iobj & 60 \\
            advmod & 30         &   cc & 70 \\
            nmod & 30           &   amod & 75 \\
            nmod:tmod & 30      &   nummod & 75 \\
            nmod:npmod & 30     &   compound & 80 \\
            nmod:poss & 30      &   compound:prt & 80 \\
            xcomp & 40          &   fixed & 80 \\
            aux & 40            &   conj-n & 90 \\
            aux:pass & 40       &   conj-vb & 90 \\
            obl & 50            &   flat & 100 \\
         \hline
    \end{tabular}
    \caption{Universal Dependency relation hierarchy. The smaller a relation's level-id is, the higher that relation is in the hierarchy.}
    \label{tab:my_label}
\end{table}

\subsection{Binarization}
To preprocess the dependency parse graph, we designed a binarization algorithm that can map each dependency tree to an s-expression \cite{reddy-etal-2016-transforming}. Formally, an s-expression has the form (exp1 exp2 exp3), where exp1 is a dependency label, and both exp2 and exp3 are either (1) a word such as \textit{eat}; or (2) an s-expression such as (\textbf{det} \textit{all dogs}). The process of mapping a dependency tree to an s-expression is called binarization. Our system represents an s-expression as a binary tree. A binary tree has a root node, a left child node, and a right child node. In representing an s-expression, the root node can either be a single word or a dependency label. Both the left and the right child nodes can either be a sub-binary-tree, or null. The system always puts the modifiers on the left and the headwords on the right. For example, the sentence \textit{All dogs eat apples} has an s-expression 
\begin{figure}
    \centering
    \includegraphics[width=6cm]{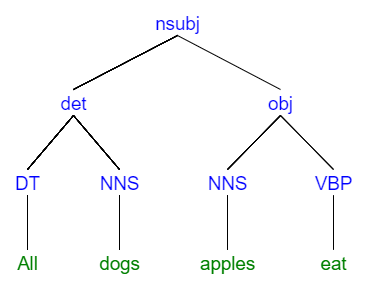}
    \caption{A binarized dependency parse tree for "All dogs eat apples."}
    \label{fig:my_label}
\end{figure}
\begin{align*}
    (\textbf{nsubj}\;  (\textbf{det}\; All\;  dogs)\; (\textbf{obj}\; eat\; apples))
\end{align*}
and can be shown as a binary tree in figure 2. In the left sub-tree (\textit{All dogs}), the dependency label \textbf{det} will be the root node, the modifier \textit{all} will be the left child, and the headword \textit{dogs} will be the right child.   

Our binarization algorithm employs a dependency relation hierarchy to impose a strict traversal order from the root relation to each leaf word. The hierarchy allows for an ordering on the different modifier words. For example, in the binary dependency parse tree (\textbf{nsubj}\;  (\textbf{det}\; All\;  dogs)\; (\textbf{obj}\; eat\; apples)), the nominal subject (\textbf{nsubj}) goes above the determiner (\textbf{det}) in the tree because \textbf{det} is lower than \textbf{nsubj} in the hierarchy.
We originally used the binarization hierarchy from \citet{reddy-etal-2016-transforming}'s work, and later extended it with additional dependency relations such as oblique nominal (\textbf{obl}) and expletive (\textbf{expl}). Table 1 shows the complete hierarchy where the level-id indicates a relation's level in the hierarchy. The smaller a relation's level-id is, the higher that relation is in the hierarchy.

\begin{algorithm}
    \small
    \caption{Binarization}
    \begin{algorithmic}[1]
    
       \State $\mathrm{root}$ $\gets$ \Call{get\_root\_node}{$\mathcal{G}$}
       \State $\calT \gets$ \Call{compose}{$\mathrm{root}$}
       \State\Return $\calT$ \\
     
     \Function{compose}{$\mathrm{node}$}:
        \State $\mathcal{C} \gets$ \Call{get\_children}{$\mathrm{node}$}
        \State $\mathcal{C}_s \gets$ \Call{sort\_by\_priority}{$\mathcal{C}$}
        \If{$\mid\mathcal{C}_s\mid$ == 0}
            \State $\mathcal{B} \gets $ \Call{BinaryDependencyTree}{}()
            \State $\mathcal{B}$.val = $\mathrm{node}$
            \State\Return $\mathcal{B}$
        \Else
            \State $\mathrm{top} \gets \mathcal{C}.\mathrm{pop}$()
            \State $\mathcal{B} \gets $ \Call{BinaryDependencyTree}{}()
            \State $\mathcal{B}$.val = \Call{relate}{$\mathrm{top}$, $\mathrm{node}$}
            \State $\mathcal{B}$.left = \Call{compose}{$\mathrm{top}$}
            \State $\mathcal{B}$.right = \Call{compose}{$\mathrm{node}$}
            \State\Return $\mathcal{B}$
        \EndIf
       \EndFunction
    \end{algorithmic}
\end{algorithm}

\subsection{Polarization}
The polarization algorithm places polarities on each node of a UD parse tree based on a lexicon of polarization rules for each dependency relation and some special words. Our polarization algorithm is similar to the algorithms surveyed by \citet{LavalleMartnez2018EquivalencesAP}. Like the algorithm of \citet{Sanchez1991StudiesON}, our algorithm computes polarity from leaves to root. One difference our algorithm has is that often, the algorithm computes polarity following a left-to-right inorder traversal (left$\longrightarrow$root$\longrightarrow$right) or a right-to-left inorder traversal (right$\longrightarrow$root$\longrightarrow$left) in additional to the top-down traversal. In our algorithm, each node's polarity depends both on its parent node and its sibling node (left side or right side), which is different from algorithms in \citet{LavalleMartnez2018EquivalencesAP}'s paper. Our algorithm is deterministic, and thus never fails.

\begin{figure}[t!]
   \centering
    \includegraphics[width=7cm]{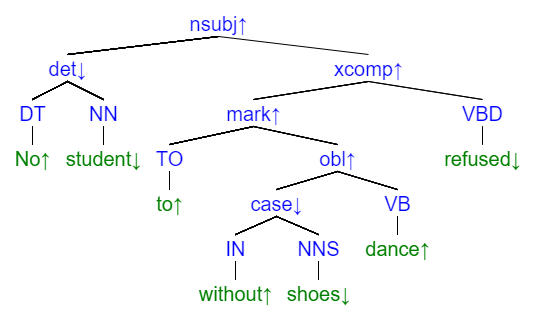}
   \caption{Visualization of a polarized binary dependency parse tree for a triple negation sentence \textit{No student refused to dance without shoes.}}
    \label{fig:exp1}
\end{figure}

 The polarization algorithm takes in a binarized UD parse tree $\calT$ and a set of polarization rules, both dependency-relation-level ($\mathcal{L}$) and word-level ($\mathcal{W}$). The algorithm outputs a polarized UD parse tree $\mathcal{T^*}$ such that (1) each node is marked with a polarity of either monotone ($\uparrow$), antitone ($\downarrow$), or no monotonicity information (=), (2) both $\calT$ and $\mathcal{T^*}$ have the same universal dependency structure except the polarity marks. Figure 3 shows a visualization of the binary dependency parse tree after polarization completes. The general steps of the polarization start from the root node of the binary parse tree. The system will get the corresponding polarization rule from the lexicon according to the root node's dependency relation. In each polarization rule, the system applies the polarization rule and then continues the above steps recursively down the left sub-tree and the right sub-tree. Each polarization rule is composed from a set of basic building blocks include rules for negation, equalization, and monotonicity generation. When the recursion reaches a leaf node, which is an individual word in a sentence, a set of word-based polarization rules will be retrieved from the lexicon, and the system polarizes the nodes according to the rule corresponding to a particular word. More details about word-based polarization rules will be covered in section 3.4.2, Polarity Generation. An overview of the polarization algorithm and a general scheme of the implementation for dependency-level polarization rules are shown in Algorithm 2.

\begin{algorithm}[t]
    \small
	\textbf{Input:} $\calT$: binary dependency tree\\
	\hspace*{2.7em} $\mathcal{L}$: dependency-level polarization rules \\
	\hspace*{2.7em} $\mathcal{W}$: word-level polarization rules \\
	\textbf{Output:} $\mathcal{T^*}$: polarized binary dependency tree\\
	
	\begin{algorithmic}[1]
	
	\If{$\calT$.is\_tree}
	    \State $\mathrm{relation} \gets \calT.\mathrm{val}$
	    \State \Call{polarization\_rule}{.} $ \gets \mathcal{L}[\mathrm{relation}]$
	    \State \Call{polarization\_rule}{$\calT$}
	\EndIf\\
	
	\AlgComment{General scheme of a polarization rule's implementation for a dependency relation}
	\Function{polarization\_rule}{$\calT$}
	    \AlgComment{Initialize or inherit polarities}
	    \If{$\calT$.mark $\neq$ NULL}
	        \State $\calT$.right.mark = $\calT$.mark
	        \State $\calT$.left.mark = $\calT$.mark
	   \Else
	        \State $\calT$.right.mark = $\uparrow$
	        \State $\calT$.left.mark = $\uparrow$
	   \EndIf \\
	   
	   \AlgComment{Polarize sub-trees}
	   \State \Call{polarization}{$\calT$.left}
	   \State \Call{polarization}{$\calT$.right}
	   
	   \AlgComment{Or, for relations like \textbf{nsubj}:}
	   \AlgComment{\Call{polarization}{$\calT$.right}}
	   \AlgComment{\Call{polarization}{$\calT$.left}} \\
	   
	   \AlgComment{Apply $\mathtt{negation}$ and $\mathtt{equalization}$ rules}
	   \If{\Call{negate}{} is applicable}
	   \State \Call{negate}{$\calT$}
	   \EndIf
	   \If{\Call{equalize}{} is applicable}
	   \State \Call{equalize}{$\calT$}
	   \EndIf\\
	   
	   \AlgComment{Apply word-level rules}
	   \If{not $\calT$.is\_tree and $\calT$.val $\in  \mathcal{W}$.keys}
	   \State \Call{word\_rule}{.} $ \gets \mathcal{W}[\calT$.val$]$
	   \State \Call{word\_rule}{$\calT$}
	   \EndIf
	   
	\EndFunction
	
	\end{algorithmic}
	
	\caption{Polarization}
\end{algorithm}

\subsection{Polarization Rules}
Our polarization algorithm contains a lexicon of polarization rules corresponding to each dependency relation. Each polarization rule is composed from a set of building blocks divided into three categories: negation rules, equalization rules, and monotonicity generation rules. The generation rules will generate three types of monotonicity: monotone ($\uparrow$), antitone ($\downarrow$), and no monotonicity information (=) either by initialization or based on the words. 
\subsubsection{Building Blocks}
\paragraph{Negation and Equalization} The negation rule and the equalization rule are used by several core dependency relations such as \textbf{nmod}, \textbf{obj}, and \textbf{acl:recl}. Both negation and equalization have two ways of application: backward or top-down. A backward negation rule is triggered by a downward polarity ($\downarrow$) on the right node of the tree (marked below as R), flipping every node's polarity under the left node (marked below as L). Similarly, a backward equalization rule is triggered by a no monotonicity information polarity (=) on the tree's right node, and it marks every node under the left node as =. Examples for trees before and after applying a backward and forward negation and equalization are shown as follows:
\begin{itemize}
    \small
    \item Backward Negation:
    
    \Tree [.$\textbf{obj}^\uparrow$ 
            $\neg(\textbf{L}^\uparrow)$
            $\textbf{R}^\downarrow$
          ]
    \Tree [.$\textbf{obj}^\uparrow$
            $\textbf{L}^\downarrow$
            $\textbf{R}^\downarrow$
          ]
    
    \item Backward Equalization:
    
    \Tree [.$\textbf{obj}^\uparrow$
            $\cong(\textbf{L}^\uparrow)$
            $\textbf{R}^{=}$
          ]
    \Tree [.$\textbf{obj}^\uparrow$
            $\textbf{L}^{=}$
            $\textbf{R}^{=}$
          ]
 
    \item Forward Negation:
    
    \Tree [.$\textbf{advmod}^\uparrow$
            $\textbf{L}^\downarrow$
            $\neg(\textbf{R}^\uparrow)$
          ]
    \Tree [.$\textbf{advmod}^\uparrow$
            $\textbf{L}^\downarrow$
            $\textbf{R}^\downarrow$
          ]
    
    \item Forward Equalization:
 
    \Tree [.$\textbf{advmod}^\uparrow$
            $\textbf{L}^{=}$
            $\cong(\textbf{R}^\uparrow)$
          ]
    \Tree [.$\textbf{advmod}^\uparrow$
            $\textbf{L}^{=}$
            $\textbf{R}^{=}$
          ]
    
\end{itemize}
where $\neg$ means negation and $\cong$ means equalization.

A top-down negation is used by the polarization rule like determiner (\textbf{det}) and adverbial modifier (\textbf{advmod}). It starts at the parent node of the current tree, and flips the arrow on each node under that parent node excluding the current tree. This top-down negation is used by \textbf{det}, \textbf{case}, and \textbf{advmod} when a negation operators like \textit{no}, \textit{not}, or \textit{at-most} appears. Below is an example of a tree before and after applying the top-down negation:
    
    {
    \small
    \center
    \Tree [.$\neg(\textbf{nsubj}^\uparrow)$
            [.$\textbf{det}^\uparrow$
              $\textbf{No}^\uparrow$
              $\textbf{cat}^\downarrow$
            ]
            $\neg(\textbf{flies}^\uparrow)$
          ]
    \Tree [.$\textbf{nsubj}^\downarrow$
            [.$\textbf{det}^\uparrow$
              $\textbf{No}^\uparrow$
            $\textbf{cat}^\downarrow$
            ]
            $\textbf{flies}^\downarrow$
          ]}

\paragraph{Polarity Generation} The polarity is generated by words. During the polarization, the polarity can change based on a particular word that can promote the polarity governing the part of the sentence to which it belongs. These words include quantifiers and verbs. For the monotonicity from quantifiers, we follow the monotonicity profiles listed in the work done by \citet{icard-iii-moss-2014-recent} on monotonicity, which built on \citet{vanBenthemEssays86}.
 Additionally, to extend to more quantifiers, we observed polarization results generated by ccg2mono. Overall, we categorized the quantifiers as follows: 

\begin{itemize}
    \small
    \item Universal Type
    \begin{align*}
    \mathrm{Every} \downarrow \;\; \uparrow \;\;\;\; \mathrm{Each} \downarrow \;\; \uparrow \;\;\;\; \mathrm{All} \downarrow \;\; \uparrow \;\;\;\;\;\;\;\;\;\;
\end{align*}
    \item Negation Type
    \begin{align*}
    \mathrm{No} \downarrow \;\; \downarrow \;\;\;\; \mathrm{Less \; than}  \downarrow \;\; \downarrow \;\;\;\; \mathrm{At \; most} \downarrow \;\; \downarrow
    \end{align*}
    \item Exact Type
    \begin{align*}
    \;\;\;\; \mathrm{Exactly \; n} = \;\; = \;\;\;\; \mathrm{The} = \;\; \uparrow \;\;\;\; \mathrm{This} = \;\; \uparrow \;\;\;\; \;\;\;\;\;
    \end{align*}
    \item Existential Type
    \begin{align*}
    \mathrm{Some} \uparrow \;\; \uparrow \;\;\;\; \mathrm{Several} \uparrow \;\; \uparrow \;\;\;\; \mathrm{A, An} \uparrow \;\; \uparrow
    \end{align*}
    \item Other Type
    \begin{align*}
    \mathrm{Most} = \;\; \uparrow \;\;\;\; \mathrm{Few} = \;\; \downarrow  \;\;\;\; \;\;\;\; \;\;\;\; \;\;\;\; \;\;\;\;
    \end{align*}
\end{itemize}
Where the first mark is the monotonicity for the first argument after the quantifier and the second mark is the monotonicity for the second argument after the quantifier.
\noindent For verbs, there are upward entailment operators and downward entailment operators. Verbs that are downward entailment operators, such as \textit{refuse}, promote an antitone polarity, which will negate its dependents. For example, for the phrase \textit{refused to go}, \textit{refused} will promote an antitone polarity, which negates \textit{to dance}:

{\small \center
\Tree [.$\textbf{xcomp}^\uparrow$
        [.$\neg(\textbf{mark}^\uparrow)$
          $\neg(\textbf{to}^\uparrow)$ 
          $\neg(\textbf{go}^\uparrow)$
        ]
        $\textbf{refused}^\uparrow$
      ] 
\Tree [.$\textbf{xcomp}^\uparrow$
        [.$\textbf{mark}^\downarrow$
         $\textbf{to}^\downarrow$ 
         $\textbf{go}^\downarrow$
        ]
        $\textbf{refused}^\uparrow$
      ] } \newline
      
\noindent In addition to quantifiers and verbs, some other words also change the monotonicity of a sentence. For example, words like \textit{not}, \textit{none}, and  \textit{nobody} promote an antitone polarity. Our system also handles material implications with the form \textit{if x then y}. Based on \citet{Moss2012TheSO}, the word \textit{if} promotes an antitone polarity in the antecedent and positive polarity in the consequent. For background on monotonicity and semantics, see  \citet{vanBenthemEssays86}, \citet{KeenanFaltz}, and also \citet{10.5555/2387636.2387659}.

\subsubsection{Dependency Relation Rules}
Each dependency relation has a corresponding polarization rule. All the rules start with initializing the starting node as upward monotone polarity ($\uparrow$). Alternatively, if the starting node has a polarity marked, each child node will inherit the root node's polarity. Each rule's core part is a combination of the default rules and monotonicity generation rules. In this section, we will briefly show three major types of dependency relation rules in the polarization algorithm. The relative clause modifier relation will represent rules for modifier relations. The determiner relation rule will represent rules containing monotonicity generation rules. The Object and open clausal complement rule will represent rules containing word-level polarization rules. 

\paragraph{Relative Clause Modifier} For the relative clause modifier relation (\textbf{acl:relcl}), the relative clause depends on the noun it modifies. First, the polarization will first be performed on both the left and right nodes, and then, depending on the polarity of the right node, a negation or an equalization rule will be applied. The algorithm first applies a top-down inheritance if the root already has its polarity marked; otherwise, it initializes the left and right nodes as monotone. The algorithm polarizes both the left and right nodes. Next, the algorithm checks the right node's polarity. If the right node is marked as antitone, a backward negation is applied. Alternatively, if the right node is marked as no monotonicity information, a backward equalization is applied. During the experiments, we noticed that if the root node is marked antitone, and the left node inherits that, a negation later will cause a double negation, producing incorrect polarity marks. To avoid this double negation, we exclude the left node from the top-down inheritance rule by initializing the left node directly with a monotone mark. The rule for \textbf{acl:relcl} also applies to the adverbial clause modifier (\textbf{advcl}) and the clausal modifier of noun (\textbf{acl}). An overview of the algorithm is shown in Algorithm 3.
\begin{algorithm}[t]
    \small
    \caption{Polarize\_acl:relcl}
    \textbf{Input:} $\calT$: binary dependency sub-tree\\
	\textbf{Output:} $\mathcal{T^*}$: polarized binary dependency sub-tree\\
    \begin{algorithmic}[1]
        \If{$\calT$.mark $\neq$ NULL}
            \State $\calT$.right.mark = $\calT$.mark
        \Else
            \State $\calT$.right.mark = $\uparrow$
        \EndIf
        \State $\calT$.left.mark = $\uparrow$\\
        
        \State \Call{polarize}{$\calT$.right}
        \State \Call{polarize}{$\calT$.left}\\
        
        \If{$\calT$.right.mark == $\downarrow$}
            \State \Call{negate}{$\calT$.left}
        \ElsIf{$\calT$.right.mark == $=$}
            \State \Call{equalize}{$\calT$.left}
        \EndIf
    \end{algorithmic}
\end{algorithm}

\begin{algorithm}[t!]
    \small
    \caption{Polarize\_det}
    \textbf{Input:} $\calT$: binary dependency sub-tree\\
    \hspace*{2.7em} $\mathcal{D}$: determiner mark dictionary \\
	\textbf{Output:} $\mathcal{T^*}$: polarized binary dependency sub-tree\\
    \begin{algorithmic}[1]
            \State $\mathrm{det\_type}$ $\gets$ \Call{get\_det\_type}{$\calT$.left}
            
            \If{$\calT$.mark $\neq$ NULL}
                \State $\calT$.left.mark = $\calT$.mark
            \Else
                \State $\calT$.left.mark = $\uparrow$
            \EndIf\\
            
            \State $\calT$.right.mark = $\mathcal{D}$[det\_type]
            \State \Call{polarize}{$\calT$.right}\\
            
            \If{det\_type == $\mathtt{negation}$}
                \State \Call{negate}{$\calT$.parent}
            \EndIf
    \end{algorithmic}
\end{algorithm}

\paragraph{Determiner}
 For the determiner relation (\textbf{det}), each different determiner can assign a new monotonicity to the noun it modifies. First, the algorithm performs a top-down inheritance on the left node if the root already has polarity marked. Next, the algorithm assigns the polarity for the noun depending on the determiner's type. For example, if the determiner is a universal quantifier, an antitone polarity is assigned to the right node. For negation quantifiers like \textit{no}, its right node also receives an antitone polarity. Thus,  a top-down negation is applied at the determiner relation tree's parent. Algorithm 4 shows an overview of the algorithm.

\begin{algorithm}[b]
    \small
    \caption{Polarize\_obj}
    \textbf{Input:} $\calT$: binary dependency sub-tree\\
	\textbf{Output:} $\mathcal{T^*}$: polarized binary dependency sub-tree\\
    \begin{algorithmic}[1]
        \If{$\calT$.mark $\neq$ NULL}
            \State $\calT$.right.mark = $\calT$.mark
        \Else
            \State $\calT$.right.mark = $\uparrow$
        \EndIf
        \State $\calT$.left.mark = $\uparrow$\\
        
        \State \Call{polarize}{$\calT$.right}
        \State \Call{polarize}{$\calT$.left}\\
        
        \AlgComment{Word-level polarization rule for downward entailment operators}
        \If{\Call{is\_downward\_operator}{$\calT$.right.mark}}
            %\State $\calT$.mark = $\downarrow$
            \State \Call{negate}{$\calT$.left}
        \EndIf\\
    \end{algorithmic}
\end{algorithm}

\paragraph{Object and Open Clausal Complement}
For the object relation (\textbf{obj}) and the open clausal complement relation \textbf{xcomp}, both the verb and the noun would inherit the monotonicity from the parent in the majority of cases. The inheritance procedure is the same as the one used in \textbf{acl:relcl}'s rule. Similarly, after the inheritance, the rule will polarize both the right sub-tree and the left sub-tree. Differently, since \textbf{obj} and \textbf{xcomp} both have a verb under the relation, they require a word-level polarization rule that will check the verb determine if the verb is a downward entailment operator, which prompts an antitone monotonicity. The algorithm takes in a dictionary that contains a list of verbs and their implicatives. The dictionary is generated from the implicative verb dataset made by \citet{ross-pavlick-2019-well}. If a verb is a downward entailment operator, which has a negative implicative, the rule will apply a negation rule on the left sub-tree to flip each node's arrow in the left sub-tree. An overview of the algorithm is shown in Algorithm 5.

%\paragraph{Nominal Subject}
%Here we use the nominal subject relation (\textbf{nsubj}) as an example to show the general flow of a dependency relation.   
%\begin{algorithm}[h]
%     \small
%   \caption{Polarize-nsubj}
%   \begin{algorithmic}[1]
%        \Function{polarize-nsubj}{tree}
%            
%            \If{tree.mark != NULL}
%                \State tree.left.mark = tree.mark
%                \State tree.right.mark = tree.mark
%            \Else
%                \State tree.left.mark = $\uparrow$
%                \State tree.right.mark = $\uparrow$
%            \EndIf
%            \State \textit{polarize}(tree.right)
%            \State \textit{polarize}(tree.left)
%            \If{tree.mark == NULL}
%                \State tree.mark = tree.right.mark
%            \EndIf
%        \EndFunction
%   \end{algorithmic}
%\end{algorithm}

\begin{table*}[t!]
    \centering
    \small
    \begin{tabular}{|c|c|}
        \hline sentence & type \\ \hline
         More$^\uparrow$ dogs$^\uparrow$ than$^\uparrow$ cats$^\downarrow$ sit$^=$ & comparative \\
         Less$^\uparrow$ than$^\uparrow$ 5$^\uparrow$ people$^\downarrow$ ran$^\downarrow$ & less-than \\
         A$^\uparrow$ dog$^\uparrow$ who$^\uparrow$ ate$^\uparrow$ two$^=$ rotten$^\uparrow$ biscuits$^\uparrow$ was$^\uparrow$ sick$^\uparrow$ for$^\uparrow$ three$^\downarrow$ days$^\downarrow$ & number\\
         Every$^\uparrow$ dog$^\downarrow$ who$^\downarrow$ likes$^\downarrow$ most$^\downarrow$ cats$^=$ was$^\uparrow$ chased$^\uparrow$ by$^\uparrow$ at$^\uparrow$ least$^\uparrow$ two$^\downarrow$ of$^\uparrow$ them$^\uparrow$ & every:most:at-least \\
         Even$^\uparrow$ if$^\uparrow$ you$^\downarrow$ are$^\downarrow$ addicted$^\downarrow$ to$^\downarrow$ cigarettes$^\downarrow$ you$^\uparrow$ can$^\uparrow$ smoke$^\uparrow$two$^\downarrow$ a$^\uparrow$ day$^\uparrow$ & conditional:number\\
         \hline
    \end{tabular}
    \caption{Example sentences in \citet{huMoss2020Tsinghua}'s evaluation dataset}
    \label{tab:my_label}
\end{table*}

\section{Comparison to Existing Systems}
We conducted several preliminary comparisons to two existing systems. First, we compared to NatLog's monotonicity annotator. Natlog's annotator also uses dependency parsing. The polarization algorithm does pattern-based matching for finding occurrences of downward monotonicity information, and the algorithm only polarizes on word-level. In contrast, our system uses a tree-based polarization algorithm that polarizes both on word-level polarities and constituent level polarities. Our intuition is that the Tregex patterns used in NatLog is not as common or as easily understandable as the binary tree structure, which is a classic data structure wildly used in the filed of computer science.  

According to the comparison on a list of sentences, NatLog's annotator does not perform as well as our system. For example, for a phrase \textit{the rabbit}, \textit{rabbit} should have a polarity with no monotonicity information (=). However, NatLog marks \textit{rabbit} as a monotone polarity ($\uparrow$). NatLog also incorrectly polarizes sentences containing multiple negations. For example, for a triple negation sentence, \textit{No newspapers did not report no bad news}, NatLog gives: \textit{No$^\uparrow$ newspapers$^\downarrow$ did$^\downarrow$ not$^\downarrow$ report$^\uparrow$ no$^\uparrow$ bad$^\uparrow$ news$^\uparrow$}. This result has incorrect polarity marks on multiple words, where \textit{report}, \textit{bad}, \textit{news} should be $\downarrow$, and \textit{no} should be $\uparrow$. Both of the scenarios above can be handled correctly by our system.  

Comparing to ccg2mono, our algorithm shares some similarities to its polarization algorithm. Both of the systems polarize on a tree structure and rely on a lexicon of rules, and they both polarize on the word-level and the constituent level. One difference is that ccg2mono's algorithm contains two steps, the first step puts markings on each node, and the second step puts polarities on each node. Our system does not require the step of adding markings and only contains the step of adding polarities on each node.

Our system has multiple advantages over ccg2mono. For parsing, our system uses UD parsing, which is more accurate than CCG parsing used by ccg2mono due to a large amount of training data. Also, our system covers more types of text than ccg2mono because UD parsing works for a variety of text genres such as web texts, emails, reviews, and even informal texts like Twitter tweets. \cite{silveira14gold, Zeldes2017, liu-etal-2018-parsing}. Our system can also work for more languages than ccg2mono since UD parsing supports more languages than CCG parsing. 

Overall, our system delivers more accurate polarization than ccg2mono. Many times the CCG parser makes mistakes and leads to polarization mistakes later on. For example, in the annotation \textit{The}$^\downarrow$ \textit{market}$^\downarrow$ \textit{is}$^\downarrow$ \textit{not}$^\downarrow$ \textit{impossible}$^\downarrow$ \textit{to}$^\downarrow$ \textit{navigate}$^\downarrow$, ccg2mono incorrectly marks every word as $\downarrow$. Our system, on the other hand, uses UD parsing which has higher parsing accuracy than CCG parsing, and thus leads to fewer polarization mistakes compared to ccg2mono. For the expression above, our system correctly polarizes it as \textit{The}$^\uparrow$ \textit{market}$^=$ \textit{is}$^\uparrow$  \textit{not}$^\uparrow$ \textit{impossible}$^\downarrow$  \textit{to}$^\uparrow$  \textit{navigate}$^\uparrow$.

Our system also handles multi-word quantifiers better than ccg2mono. For example, for a multi-word quantifier expression like \textit{all of the dogs}, ccg2mono mistakenly marks \textit{dogs} as $=$. Our system, however, can correctly mark the expression: \textit{all}$^\uparrow$  \textit{of}$^\uparrow$ \textit{the}$^\uparrow$ \textit{dogs}$^\downarrow$.

Moreover, the core of ccg2mono does not include aspects of verbal semantics of downward-entailing operators like \textit{forgot} and \textit{regret} \cite{MossHuSoundness}. For example ccg2mono's polarization for \textit{Every}$^\uparrow$ \textit{member}$^\downarrow$ \textit{forgot}$^\uparrow$ \textit{to}$^\uparrow$ \textit{attend}$^\uparrow$ \textit{the}$^\uparrow$ \textit{meeting}$^=$ is not correct because it fails to flip the polarity of \textit{to attend the}. In contrast, our system produces a correct result: \textit{Every}$^\uparrow$ \textit{member}$^\downarrow$ \textit{forgot}$^\uparrow$ \textit{to}$^\downarrow$ \textit{attend}$^\downarrow$ \textit{the}$^\downarrow$ \textit{meeting}$^=$.

All three systems have difficulty polarizing sentences containing numbers. A scalar number \textbf{n}'s monotonicity information is hard to determine because it can presenter different contexts: a single number \textbf{n}, without additional quantifiers or adjectives, can either mean \textit{at least} \textbf{n}, \textit{at most} \textbf{n}, \textit{exactly} \textbf{n}, and \textit{around} \textbf{n}. These contexts are syntactically hard to identify for a dependency parser or a CCG parser because it would require pragmatics and some background knowledge which the parsers do not have. For example, in the sentence \textit{A dog ate 2 rotten biscuits}, the gold label for \textit{2} is $=$ which indicates that the context is "exactly 2". However, our system marks this as "$\downarrow$ since it considers the context as "at least 2", which is different from the gold label.

\section{Experiment}
\paragraph{Dataset} We obtained the small evaluation dataset used in the evaluation of ccg2mono \cite{huMoss2020Tsinghua} from its authors. The dataset contains 56 hand-crafted English sentences, each with manually annotated monotonicity information. The sentences cover a wide range of linguistic phenomena such as quantifiers, conditionals, conjunctions, and disjunctions. The dataset also contains hard sentences involving scalar numbers. Some example sentences from the dataset are shown in Table 2.

\paragraph{Dependency Parser} In order to obtain a universal dependency parse tree from a sentence, we utilize a parser from Stanza \cite{qi-etal-2020-stanza}, a Python natural language analysis package made by Stanford. The neural pipeline in Stanza allow us to use pretrained neural parsing models to generate universal dependency parse trees. To achieve optimal performance, we trained two neural parsing models: one parsing model trained on Universal Dependency English GUM corpus \cite{Zeldes2017}. The pretrained parsing model achieved 90.0 LAS \cite{zeman-etal-2018-conll} evaluation score on the testing data.

\paragraph{Experiment Setup} We evaluated the polarization accuracy on both the token level and the sentence level, in a similar fashion to the evaluation for part-of-speech tagging \cite{Manning2011PartofSpeechTF}. For both levels of accuracy, we conducted one evaluation on all tokens (\textit{acc(all-tokens)} in Table 3) and another one on key tokens including content words (nouns, verbs, adjectives, adverbs), determiners, and numbers (\textit{acc(key-tokens)} in Table 3). The key tokens contain most of the useful monotonicity information for inference. In token-level evaluation, we counted the number of correctly annotated tokens for \textit{acc(all-tokens)} or the number of correctly annotated key tokens for \textit{acc(key-tokens)}. In sentence-level evaluation, we counted the number of correct sentences. A correct sentence has  all tokens correctly annotated for \textit{acc(all-tokens)} or all key tokens correctly annotated for \textit{acc(key-tokens)}. We also evaluated our system's robustness on the token level. We followed the robustness metric for evaluating multi-class classification tasks, which uses precision, recall, and F1 score to measure a system's robustness. We calculated these three metrics for each polarity label: monotone($\uparrow$), antitone($\downarrow$), and None or no monotonicity information($=$). The robustness evaluation is also done both on all tokens and on key tokens. 

\begin{table}[t]
\footnotesize
\centering
\begin{tabular}{|c||c|c|c|}
\hline  & \multicolumn{3}{|c|}{\textbf{Token-level}}\\ \hline 
system  & NatLog & ccg2mono & ours \\\hline
acc(all-tokens) & 69.9 & 76.0 & \textbf{96.5}\\
acc(key-tokens) & 68.1 & 78.0 & \textbf{96.5}\\
\hline  & \multicolumn{3}{|c|}{\textbf{Sentence-level}} \\ \hline
system  & NatLog & ccg2mono & ours \\\hline
acc(all-tokens) & 28.0 & 44.6 & \textbf{87.5}\\
acc(key-tokens) & 28.6 & 50.0 & \textbf{89.2}\\
\hline
\end{tabular}
\caption{\label{font-table} This table shows the polarity annotation accuracy on the token level and the sentence level for three systems: NatLog, ccg2mono, and our system. The token level accuracy counts the number of correctly annotated tokens, and the sentence level accuracy counts the number of correctly annotated sentences. Two types of accuracy are used. For \textit{acc(all-tokens)}, all tokens are evaluated. For \textit{acc(key-tokens)}, only key tokens (content words + determiners + numbers) are evaluated.}
\end{table}

\begin{table*}[t!]
\footnotesize
\centering
\begin{tabular}{|c||c|c|c|c|c|c|c|c|c|}
\hline  & \multicolumn{9}{|c|}{\textbf{All Tokens}} \\ \hline 
system & \multicolumn{3}{|c|}{\textbf{NatLog}} & \multicolumn{3}{|c|}{\textbf{ccg2mono}} & \multicolumn{3}{|c|}{\textbf{ours}} \\ \hline
Polarity  & Monotone & Antitone & None  & Monotone & Antitone & None  & Monotone & Antitone & None \\ \hline
precision & 71.4 & 43.5 & 70.7 & 86.0 & 75.6 & 58.0 & \textbf{97.6} & \textbf{96.5} & \textbf{91.7} \\
recall & 87.3 & 15.9 & 63.9 & 77.8 & 78.3 & 74.6 & \textbf{97.2} & \textbf{89.4} & \textbf{87.3} \\
F1-score & 78.6 & 23.3 & 67.1 & 81.7 & 76.9 & 65.3 & \textbf{97.4} & \textbf{97.6} & \textbf{89.4} \\
\hline  & \multicolumn{9}{|c|}{\textbf{Key Tokens}} \\ \hline
system & \multicolumn{3}{|c|}{\textbf{NatLog}} & \multicolumn{3}{|c|}{\textbf{ccg2mono}} & \multicolumn{3}{|c|}{\textbf{ours}} \\ \hline
Polarity  & Monotone & Antitone & None  & Monotone & Antitone & None  & Monotone & Antitone & None \\ \hline
precision & 68.7 & 70.9 & 42.1 & 85.2 & 78.7 & 62.7 & \textbf{96.9} & \textbf{96.4} & \textbf{94.2} \\
recall & 88.6 & 61.5 & 14.0 & 80.3 & 79.3 & 73.7 & \textbf{97.9} & \textbf{98.5} & \textbf{86.0} \\
F1-score & 77.4 & 65.9 & 21.1 & 82.7 & 79.0 & 67.7 & \textbf{97.4} & \textbf{97.4} & \textbf{89.9} \\

\hline
\end{tabular}
\caption{\label{font-table} Token level robustness comparison between NatLog, ccg2mono, and our system. The robustness score is evaluated both on all tokens and on key tokens (content words + determiners + numbers). For each of the three polarities: monotone($\uparrow$), antitone($\downarrow$), and None or no monotonicity information($=$), the relative precision, recall and F1 score are calculated.}
\end{table*}

\section{Evaluation} Table 3 shows the performance of our system, compared with NatLog and ccg2mono. Our evaluation process is the same as \citet{huMoss2020Tsinghua}. From Table 3, we first observe that our system consistently outperforms ccg2mono and NatLog on both the token level and the sentence level. For accuracy on the token level, our system has the highest accuracy for the evaluation on all tokens (96.5) and the highest accuracy for the evaluation on key tokens (96.5). Our system's accuracy on key tokens is higher than the accuracy on all tokens, which demonstrates our system's good performance on polarity annotation for tokens that are more significant to monotonicity inference. For accuracy on the sentence level, our system again has the highest accuracy for the evaluation on all tokens (87.5) and the highest accuracy for the evaluation on key tokens (89.2). Such results suggest that our system can achieve good performance on determining the monotonicity of the sentence constituents. Overall, the evaluation validates that our system has higher polarity annotation accuracy than existing systems. We compared our annotations to ccg2mono's annotation and observed that of all the tokens in the 56 sentences, if ccg2mono annotates it correctly, then our system also does so. This means, our system's polarization covers more linguistic phenomena than ccg2mono. Table 4 shows the robustness score of our system and the two existing systems. Our systems has much higher precision and recall on all three polarity labels than the other two systems. For the F1 score, our system again has the highest points over the other two systems. The consistent and high robustness scores show that our system's  performance is much more robust on the given dataset than existing systems.  
 
\section{Conclusion and Future Work}
In this paper, we have demonstrated our system's ability to automatically annotate monotonicity information (polarity) for a sentence by conducting polarization on a universal dependency parse tree. The system operates by first converting the parse tree to a binary parse tree and then marking polarity on each node according to a lexicon of polarization rules. The system produces accurate annotations on sentences involving many different linguistic phenomena such as quantifiers, double negation, relative clauses, and conditionals. Our system had better performance on polarity marking than existing systems including ccg2mono \cite{hu-moss-2018-polarity} and NatLog \cite{maccartney-manning-2009-extended, angeli-etal-2016-combining}. Additionally, by using UD parsing, our system offers many advantages. Our system supports a variety of text genres and can be applied to many languages. In general, this paper opens up a new framework for performing inference, semantics, and automated reasoning over UD representations.

For future work, an inference system can be made that utilizes the monotonicity information annotated by our system, which is similar to the MonaLog system \cite{monalog}. Several improvements can be made to the system to obtain more accurate annotations. One improvement would be to incorporate pragmatics to help determine the monotonicity of a scalar number.

\section*{Acknowledgements}
This research is advised by Dr. Lawrence Moss from Indiana University and Dr. Michael Wollowski from Rose-hulman Institute of Technology. We thank their helpful advises and feedback on this research. We also thank the anonymous reviewers for their insightful comments. 

\bibliography{anthology,custom}
\bibliographystyle{acl_natbib}

%\clearpage
%\appendix
%\section{Additional Polarization Rules}
%\section{Error Analysis}

\end{document}